\documentclass[preprint,12pt]{elsarticle}

\usepackage{amssymb}
\usepackage{amsmath}

\usepackage{amssymb, xcolor, hyperref, bm, subcaption, caption, multirow, algpseudocode, algorithm, multirow, amsmath, makecell, lscape, placeins, booktabs}

\usepackage{pgfplots}

\DeclareMathOperator*{\precision}{Precision}
\DeclareMathOperator*{\recall}{Recall}
\DeclareMathOperator*{\fscore}{F1-Score}
\newcommand{\mli}[1]{\mathit{#1}}

\journal{International Journal of Human-Computer Studies}

\begin{document}

\begin{frontmatter}



\title{Towards a Novel Measure of User Trust in XAI Systems}



\author[1,2,6]{Miquel Mir\'{o}-Nicolau} \corref{cor1}
\ead{miquel.miro@uib.es}
\author[1,2,6]{Gabriel Moy\`{a}-Alcover} 
\ead{gabriel.moya@uib.es}
\author[1,2,6]{Antoni Jaume-i-Cap\'{o}}
\ead{antoni.jaume@uib.es}
\author[2,4,5,6]{Manuel Gonz\'{a}lez-Hidalgo} 
\ead{manuel.gonzalez@uib.es}
\author[7]{Adel Ghazel} 
\ead{adel.ghazel@esigelec.fr}
\author[3]{Maria Gemma Sempere Campello} 
\ead{gemma.sempere@ssib.es}
\author[3]{Juan Antonio Palmer Sancho} 
\ead{joan.palmer@ssib.es}

\cortext[cor1]{Corresponding author}
\address[1]{UGiVIA Research Group, University of the Balearic Islands, Dpt. of Mathematics and Computer Science, 07122 Palma (Spain)}
\address[2]{Laboratory for Artificial Intelligence Applications (LAIA@UIB), University of the Balearic Islands, Dpt. of Mathematics and Computer Science, 07122 Palma (Spain)}
\address[3]{Hospital Universitari Son Espases, 07010 Palma (Spain)}
\address[4]{SCOPIA Research Group, Department of Mathematical Sciences and Computer Science, University of the Balearic Islands, 07122 Palma (Spain)}
\address[5]{Institute for Health Research of the Balearic Islands (IdISBa), 07010 Palma (Spain)}
\address[6]{Artificial Intelligence Research Institute of the Balearic Islands (IAIB), University of the Balearic Islands, 07010 Palma (Spain)}
\address[7]{Univ Rouen Normandie, ESIGELEC, Normandie Univ, IRSEEM UR 4353, F-76000 Rouen (France)}

\begin{abstract}
The increasing reliance on Deep Learning models, combined with their inherent lack of transparency, has spurred the development of a novel field of study known as eXplainable AI (XAI) methods. These methods seek to enhance the trust of end-users in automated systems by providing insights into the rationale behind their decisions. This paper presents a novel trust measure in XAI systems, allowing their refinement. Our proposed metric combines both performance metrics and trust indicators from an objective perspective. To validate this novel methodology, we conducted three case studies showing an improvement respect the state-of-the-art, with an increased sensitiviy to different scenarios.

\end{abstract}


\begin{highlights}
    \item Introduced four trust-performance hybrid measures: Trust True (TT), Untrust True (UT), Trust False (TF), and Untrust False (UF), forming a confusion-matrix-like structure.
    \item Improved upon existing trust measures by incorporating model correctness and penalizing blind trust or mistrust more effectively.
    \item Demonstrated the metric's utility across three case studies, including hypothetical and real-world medical AI settings.
    \item Revealed inter-user variability in trust, even among experts, reinforcing the user-dependent nature of trust in XAI.
    \item Validated findings through qualitative user feedback and post-study questionnaires, aligning behavioral trust scores with subjective insights.
    \item Identified low trust levels despite high model accuracy in medical image analysis, highlighting issues with interpretability and explanation clarity.
\end{highlights}


\end{frontmatter}



\section{Introduction}

From the seminal work of Krizhevsky \emph{et al.}~\cite{krizhevsky2012imagenet} in 2012, machine learning models, and in particular Deep Learning (DL) ones, have become pervasive in multiple and diverse study fields. This ubiquity of DL approaches is provoked due to their far better results in comparison to the non-deep learning methods. The improvement of these methods is obtained to their high complexity, however this high complexity have provoked an increased difficulty on the understanding of their inner working. The fact that the causes behind a decision are unknown can be ignored in non-sensitive fields, nonetheless is crucial in sensitive fields as medical related task \cite{miro2022evaluating}. 

To address this issue, eXplainable Artificial Intelligence (XAI) emerge, according to Adadi and Berrada, aiming to “create a suite of techniques that produce more explainable models whilst maintaining high performance levels” ~\cite{Adadi2018}. The growing dynamic around XAI has been reflected in several scientific events and in the increase of publications as indicated in several recent reviews about the topic (\cite{Adadi2018, dosilovic_explainable_2018, murdoch_definitions_2019, anjomshoae_explainable_2019, minh_explainable_2022, barredo_arrieta_explainable_2020}). In particular, its importance is crucial in sensitive field of health and well-being (\cite{eitel2019testing, miro2022evaluating, VanderVelden2021, chaddad_survey_2023}).

Miller~\cite{miller2019explanation} identified the need to measure different aspects of the explanations to be able to make an objective evaluation, “most of the research and practice in this area seems to use the researchers’ intuitions of what constitutes a ‘good’ explanation”. With this paper, the author started a trend to measure different aspects of the explanation and the usage of social science knowledge to make objective evaluation for XAI techniques.

Multiple aspects of an explanation can be evaluated. Bodria \emph{et al.}~\cite{bodria2023benchmarking} propose a distinction between quantitative and qualitative evaluation. Similarly, Amengual-Alcover et al.\cite{ammengual2025} categorize XAI aspects into machine-centred (quantitative) and human-centred (qualitative) dimensions. { Machine-centred aspects refer to properties that are independent of the user and can thus be assessed through algorithmic or computational methods. In contrast, human-centred aspects depend on users' perceptions and interactions with the XAI system, requiring evaluation through user studies or other qualitative approaches.} Nauta \emph{et al.}~\cite{nauta2023anecdotal} reviewed the state-of-the-art of XAI evaluation and also identify as a main issue of the field the discussion between machine-centred analysis and user centred ones. Nauta \emph{et al.}~\cite{nauta2023anecdotal} identified 12 distinct elements that can be used to evaluate explanations. For a comprehensive analysis of these evaluation features, we refer the reader to their work~\cite{nauta2023anecdotal}. Doshi-Velez and Kim~\cite{doshi2017towards}, also make a similar distinction, these authors propose a taxonomy for XAI evaluation approaches, dividing them into: Application-Grounded Evaluation, Human-Grounded Metrics and Functionally-Grounded Evaluation. With the latter including machine centred approaches and the first two human-centred measures. Vilone and Long~\cite{vilone2021notions} also reviews the state-of-the-art and proposed to divide XAI measures between human centred and objective evaluation. { As can be seen, all these authors highlight the primary distinction between XAI aspects as whether they are machine-centred or human-centred.}

While machine-centred aspects are laregly studied, human-centred approaches need to further research. Barredo-Arrieta \emph{et al.}~\cite{barredo_arrieta_explainable_2020}, after reviewing the state-of-the-art, identified trust as one of the primary goals of an XAI model from a user point of view. According to Miller~\cite{miller2019explanation}, trust must be prioritised and used as a basic criterion of the explanation correctness. We followed the definition of trust from Mayer \emph{et al.}~\cite{mayer1995integrative}: ``the willingness of a party to be vulnerable to the actions of another party based on the expectation that the other will perform a particular action important to the trustor, irrespective of the ability to monitor or control that other party”

Trust is not only studied in the XAI context but in general in the whole automation field (\cite{Hoffman2018}, \cite{Lee&See2004}, \cite{Adams2003}, \cite{mercado_intelligent_2016}). The author discussed that the knowledge from this field must be used in XAI. Hoffman \emph{et al.}~\cite{Hoffman2018} studies the state-of-the-art for trust measures and detects that “the scientific literature on trust (generally) presents a number of scales for measuring trust”. These scales were build with multiple questions with the goal to measure different dimensions of trust. Hoffman \emph{et al.}~\cite{Hoffman2018} identified multiple scales used in the state-of-the-art and summarise them into four questionnaires: Jian \emph{et al.}~\cite{jian_foundations_2010}, Cahour and Forzy~\cite{cahour_does_2009}, Merrit~\cite{merritt_affective_2011}, and Wang \emph{et al.}~\cite{wang_trust_2009}. 

Even so, trust is a subjective feature that depends on each user, objective evaluation of it can be made. Mohseni \emph{et al.}~\cite{mohseni_multidisciplinary_2021} identified scales and interview as subjective measurements. According to Scharowski \emph{et al.}~\cite{scharowski_trust_2022}, these subjective measures handle the attitudinal (subjective) perception of an agent for a system, criticizing the usage of questionnaires: ``the data collected using survey scales is inherently subjective, given that it reflects participants’ own perspectives". These authors proposed to make a shift on the measurement of trust from an attitudinal approach to a behavioural approach, in other words from a subjective approach to an objective one. One examples of these shift is the work from Lai \& Tan~\cite{lai_human_2019}. These authors proposed behavioural measurement of trust, defining the trust levels as the percentage of times that the end-user relies on the prediction and explanation. The authors also identified that their measure was affected by the prediction performance: ``We find that humans tend to trust correct machine predictions more than incorrect ones, which suggests that humans can somewhat effectively identify cases where machines are wrong". However, due to the simplicity of their approach, they cannot make further analysis or to get more intricate measurements. 

The influence of the performance of the AI model into the trust with it is largely studied in the state-of-the-art. Particularly, Glikson \& Woolley~\cite{glikson2020human} reviewed the literature of trust and identified that trust is ``prone to change based on the behavior of the trustes agent''. Another concept addressed by Glikson \& Woolley~\cite{glikson2020human} was the so-called trust trajectory: how trust changes when an interaction between the model and the user exist. The main conclusion was that multiple authors \cite{de2012world, dietvorst2015algorithm, manzey2012human} identified that high initial trust in the AI system tends to decrease as a result of erroneous AI outcomes. Therefore, it is clear that exist a relation between AI performance and the trust of the user in it.

The existing state-of-the-art conclude both that an objective evaluation of trust is needed and that AI performance has a large effect on trust. Consequently, in this study, we proposed a novel method for a behaviour measure of user trust in an automated system that combines both the prediction performance and the trust of the system, in a simplified and objective approach. Our proposal is based on the work from Lai \& Tan~\cite{lai_human_2019}, as one of the first behavioural approaches to measure trust. We proposed a new set of measures based on the well-established classification measures, true positives (TP), true negatives (TN), false positives (FP), and false negatives (FN), in which we incorporate trust information within each measure. We carried out a case of study in the medical context to test the proposed measurement.

The rest of this paper is organised as follows. In the next section, we describe our proposed measure for user trust in an automatic system. In Section \ref{sec:experimental_setup} we defined a set of case of studies to verify the proposed measure. Finally, in Section~\ref{sec:conclusion} we present the conclusions of this study.

\section{Trust measure}

 The goal of this paper is to propose a novel measure for user trust in a XAI system. Knowing that user trust is highly dependent on the model behaviour, we propose a novel approach based in the existing relation between the correct prediction and the trust of the user on the system.  Muir and Moray~\cite{muir_trust_1996} state that, “Results showed that operators' subjective ratings of trust in the automation were based mainly upon their perception of its competence. Trust was significantly reduced by any sign of incompetence in the automation, even one which had no effect on overall system performance.” This conclusion is shared with other authors~\cite{de2012world, dietvorst2015algorithm, manzey2012human}, as we showed in the previous section. 
 
 In the context of classification tasks, the prediction results undergo evaluation using a set of established measures, the amount of TPs, FPs, FNs, and TNs. These fundamental measures serve as the foundation for computing a variety of more intricate metrics, allowing for a comprehensive and objective analysis of performance across various dimensions. We can simplify these four metrics in a more simple binary measure: true predictions (True Positives and True Negatives) or false predictions (False Positives and False Negatives). On the other hand, the trust of a user in a system is subjective, different users can have a completely different trust on the same system, however, as discussed in by Scharowski \emph{et al.}~\cite{scharowski_trust_2022}, we can objectively measure this subjective feature.  We proposed to, from the work of Lai \& Tan~\cite{lai_human_2019}, combine the information of performance into a behavioural questioning of the user: analysing whether the user will employ or not the system for a particular sample, taking into account if the sample is correctly classified or not. We proposed four different measures that combine both the information of the trust and the correctness of the prediction:

\begin{itemize}
    \item \textbf{Trust True (TT)} prediction. The amount of correct prediction and that the user trust the corresponding explanation.
    \item \textbf{Untrust True (UT)} prediction. The amount of correct prediction and that the user did not trust the corresponding explanation.
    \item \textbf{Trust False (TF)} prediction. The amount of incorrect prediction and that the user trust the corresponding explanation.
    \item \textbf{Untrust False (UF)} prediction. The amount of incorrect prediction and that the user did not trust the corresponding explanation.
\end{itemize}

The previous four measures are calculated, counting the amount of times each case happens. These measures are summarised in Table \ref{tab:basic_measures}

\begin{table}[!htb]
\centering
\begin{tabular}{c|c|c}
Trust / Prediction & False prediction   & True prediction \\ \hline
No                 & \textbf{UF}                  & \textbf{UT}                \\ \hline
Yes                & \textbf{TF}                  & \textbf{TT}               
\end{tabular}
\caption{\label{tab:basic_measures}Basic metrics proposed in a confusion matrix format.}
\end{table}

These four measures, similarly to the classification metrics, can be combined, obtaining multiple higher level measures. Due to the clear relation between these measures and the classification ones, we consider straightforward the adaptation of any classification measures. For example, we can adapt these two metrics: 

\begin{itemize}
    \item \textbf{Precision}. Precision, in the trust context, is the proportion of TT among the total of True prediction. See equation \ref{eq:tp} for more details.
    \item \textbf{Recall}. Recall, in the trust context, is the fraction of TT among the total of trusted predictions. See equation \ref{eq:tr} for more details.
\end{itemize}

\begin{equation}
    \precision = \frac{\mli{TT}}{\mli{TT} + \mli{UT}}.
    \label{eq:tp}
\end{equation}

\begin{equation}
    \recall = \frac{\mli{TT}}{\mli{TT} + \mli{TF}}.
    \label{eq:tr}
\end{equation}

As can be seen, both metrics represent the same as in the classification context. However, both measures isolated are not enough to depict whether a user really trust a system. For example, the Recall will be 1, the maximum value, in the case the user only trust one correct prediction, untrusting the rest of samples (both correct and incorrect). However, in this hypothetical case, the Precision is going to have a very low value. For this reason it is necessary, once again as in the classification context, to combine them. To do it we propose to use the harmonic mean, known as \textbf{F1-Score}. This metric can be seen on equation \ref{eq:f1}.

\begin{equation}
    \fscore = 2 \cdot \frac{\precision \cdot \recall}{\recall + \precision}.
    \label{eq:f1}
\end{equation}

The result of this last measure, it is inside a range of $[0, 1]$. The result is easily interpretable, with a value of $1$ indicating a perfect results. Therefore, our proposal takes into account both the performance of the model and the user trust on the explanation, using already known and defined measures.

In the following section we show a set of case studies. Their goal is to show the expresivity of our approach, working in complete different contexts, with different trust and performance levels.

\section{Case of study}
\label{sec:experimental_setup}

In the previous section, we introduced a novel trust measure that leverages performance data through the use of a confusion matrix. Our goal in this section is to demonstrate how this approach behaves across a range of diverse scenarios. Specifically, we define three distinct case of study, each characterized by varying levels of trust and performance. Importantly our goal with these scenarios was not to obtain perfect performance or perfect trust, but to verify that our approach allow the detection of different behaviours. 

First, we defined a set of hypothetical results covering extreme cases. Second, we applied our approach to real results obtained from a state-of-the-art machine learning study by Petrović \emph{et al.}~\cite{PETROVIC2020104027}. Finally, we tested our method on a novel machine learning model specifically designed to assess the trust placed in medical experts. In all three cases of study we compared our proposal to the one of Lai \& Tan~\cite{lai_human_2019}: the proportion of trusted samples.

\subsection*{Case of study 1: hypotehtical trust, hypotehtical machine learning}

In this first case of study we propose three extreme scenarios. {We referred to these scenario as users because they represent extreme user behaviour related to trust.} These three extreme cases were the following:

\begin{itemize}
    \item \textbf{Perfect system user}. This first case depicts a user trusts the correct predictions and do not trust the incorrect ones.
    
    \item \textbf{Entrusted user}. This case illustrates a user who consistently places unwavering trust in any prediction. 
    
    \item \textbf{Suspicious user}. This case showed the results of a user that never trust the outcome of the AI model. 
\end{itemize}

Once defined these different users, we calculated the values of our proposed measures. For all three cases we designed we have different trust levels with the same hypothetical model performance: half of the 100 test samples were correctly classified. The existence of both correct and incorrect classification allows us to identify the behaviour of the proposed measures with completely different samples.

The results of this hypothetical scenarios can be seen in Table \ref{tab:res_cs}. The fact that performance remains unchanged while our measure varies according to the trust value demonstrates the expressiveness of our approach in capturing the user's trust in the AI system.

\begin{table}
    \centering
    \begin{tabular}{lccc}
        \toprule
        Metric          & Perfect System User   &  Entrusted User           & Suspicious User        \\ \midrule
        TT              & 50                    &  50                        & 0                           \\
        UF              & 50                    &  0                         & 50                           \\
        UT              & 0                     &  0                         & 50                           \\
        TF              & 0                     &  50                        & 0                           \\
        Precision       & 1                     &  1                         & 0                           \\
        Recall          & 1                     &  0.5                       & 0                           \\
        F1-Score        & 1                     &  0.66                      & 0                           \\ \midrule        
        Lai \& Tan~\cite{lai_human_2019}   & 0.5   & 1 & 0             \\ \bottomrule
    \end{tabular}
    \caption{Results obtained in all hypothetical scenarios.}\label{tab:res_cs}
\end{table}

These findings demonstrate the effectiveness of the proposed metrics in capturing trust levels across distinct contexts. In the first scenario, all metrics achieved perfect scores, indicating optimal performance. In the second scenario, Precision attained a perfect score, while the remaining metrics produced significantly lower values, highlighting differences in evaluation perspectives. Finally, in the third scenario, all metrics yielded the worst possible outcomes, further validating their sensitivity to varying conditions.

{ The results of our trust measure contrast with the approach proposed by Lai \& Tan~\cite{lai_human_2019}, particularly in the evaluation of the first two users. Our method considers a lack of trust in incorrect predictions to be a desirable behavior, as demonstrated in the first scenario. In contrast, Lai \& Tan~\cite{lai_human_2019} penalize this behavior. For the Entrusted User, the opposite occurs: while our approach identifies this behavior as incorrect, Lai \& Tan's method considers it valid. This highlights a key limitation that our measure aims to address—distrust in incorrect predictions should not be penalized, but rather recognized as appropriate and even desirable.}

\subsection*{Case of study 2: hypotehtical trust, real machine learning}

\begin{figure}[!htb]
	\centering
     	\subfloat[]{\includegraphics[width=0.3\textwidth]{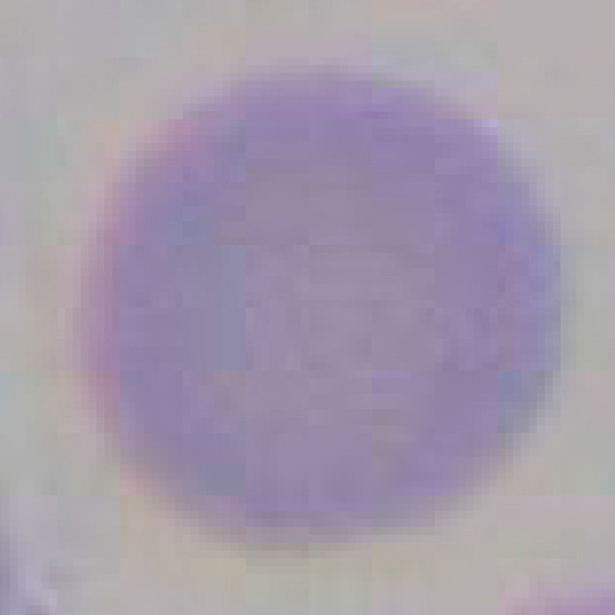}%
    	\label{subfigure:pet_circ}}
        \hfil
        \subfloat[]{\includegraphics[width=0.3\textwidth]{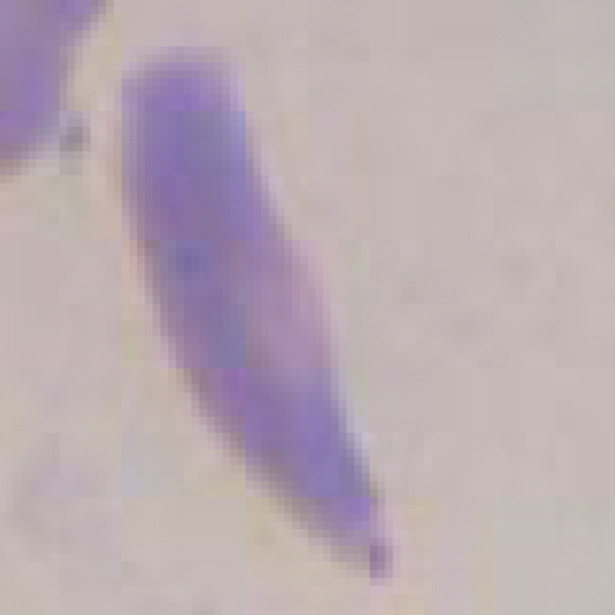}%
    	\label{subfigure:pet_elong}}
        \hfil
        \subfloat[]{\includegraphics[width=0.3\textwidth]{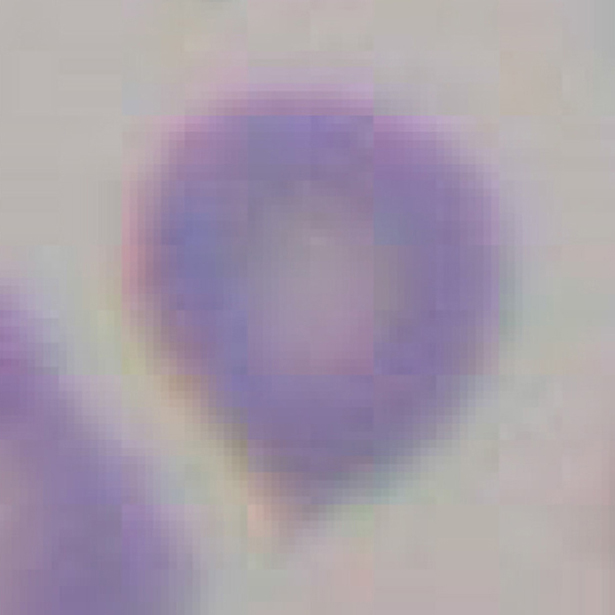}%
    	\label{subfigure:pet_other}}
    	\\
 	\caption{Samples from Petrović \emph{et al.}~\cite{PETROVIC2020104027} dataset. Each subfigure is from a different classes: circular (\ref{subfigure:pet_circ}), elongated (\ref{subfigure:pet_elong}), and other (\ref{subfigure:pet_other})}
	\label{fig:petrovic}
\end{figure}

In this second case of study we used the proposed trust measures with a real machine learning model. Particularly, we used the results from Petrović \emph{et al.}~\cite{PETROVIC2020104027}. These authors proposed a novel approach to select and train an AI model to identify and classify peripheral blood smear images of red blood cells, depending on their morphology. { Specifically, their classification problem includes three categories: elongated, circular, and others. Figure~\ref{fig:petrovic} depicts examples from each class. While the authors compared various machine learning algorithms, we selected their best-performing method, Gradient Boosting, to evaluate our trust measure. The performance of this model can be seen in the confusion matrix depictes in Table~\ref{tab:conf_mat_pet}.}

\begin{table}
    \centering
    \begin{tabular}{lccc}
        \toprule
                        & Circular      &  Elongated        & Other          \\ \midrule
        Circular        & 488           &  6                & 5              \\
        Elongated       & 8             &  194              & 8              \\
        Other           & 20            &  5                & 75             \\ \bottomrule
    \end{tabular}
    \caption{Confusion matrix of the Gradient Boosting method proposed by Petrović \emph{et al.}~\cite{PETROVIC2020104027}.}\label{tab:conf_mat_pet}
\end{table}

We recalculated and repeated the three hypothetical trust levels using the results obtained from this model. { This approach allowed us to analyse different models under the same user trust assumptions, enabling us to assess whether our trust measures are also sensitive to model performance. For simplicity, we did not distinguish between classes: a correct prediction was considered valid regardless of the predicted class. However, our approach can easily be extended to enable class-specific analyses.}

The resulting trust metrics values can be seen in Table \ref{tab:res_cs_np}. From this table we can see that the sole difference between the first case of study and this second one was the entrusted user. This case demonstrated that using an AI model with high performance values hid the bad trust results; however, the confusion matrix can be utilised to detect this unwanted behaviour. Comparing our approach to the one from Lai \& Tan~\cite{lai_human_2019}, we can see, once again, that according to their approch the users have perfect results, while clearly the trust behaviour is not the desired one.

\begin{table}
    \centering
    \begin{tabular}{lccc}
        \toprule
        Metric          & Perfect System User   &  Entrusted User       & Suspicious User        \\ \midrule
        TT              & 757                   &  757                  & 0                           \\
        UF              & 52                    &  0                    & 50                           \\
        UT              & 0                     &  0                    & 757                           \\
        TF              & 0                     &  52                   & 0                           \\
        Precision       & 1                     &  1                    & 0                           \\
        Recall          & 1                     &  0.9357               & 0                           \\
        F1-Score        & 1                     &  0.9667               & 0                           \\ \midrule        
        Lai \& Tan~\cite{lai_human_2019}   & 0.935   & 1 & 0             \\ \bottomrule
    \end{tabular}
    \caption{Results obtained in all hypothetical scenarios from the results obtained by Petrović \emph{et al.}~\cite{PETROVIC2020104027}.}\label{tab:res_cs_np}
\end{table}

These hypothetical results showed the ability of our proposal to identify different trust behaviours. Additionally, the goal of this section is to define a set of known behaviours, allowing us to compare future results with a baseline. Therefore, we used the knowledge obtained in this section to make the analysis of the results from a real study case. In the following section, we tested our measures with a real AI model.

\subsection*{Case of study 3: real trust, real machine learning}

In this third case of study we tested the proposed trust measures in a real scenario: using a real model with and tested with expert users. It is important to mention that the goal of this case of study, similarly to the previous ones, is to test the utility of our proposed measures, not to develop an XAI model. 

We investigate how our proposed measures can assess the trust of medical doctors on a real XAI approach to detect pneumonia provoked by COVID-19 from x-ray images. Particularly, from X-ray images, doctors can evaluate if there is a pulmonary involvement and its extent. In addition, they can provide a diagnosis of probability. However, in this case, the only lung disease present in the data set was COVID-19, which allowed us to classify any lung involvement as this specific disease. In this section we first introduce the main elements of the pipeline (dataset, AI model, XAI method and User recollection strategy) and we showed the case of study results.

\begin{figure}[!htb]
	\centering
     	\subfloat[]{\includegraphics[width=0.2\textwidth]{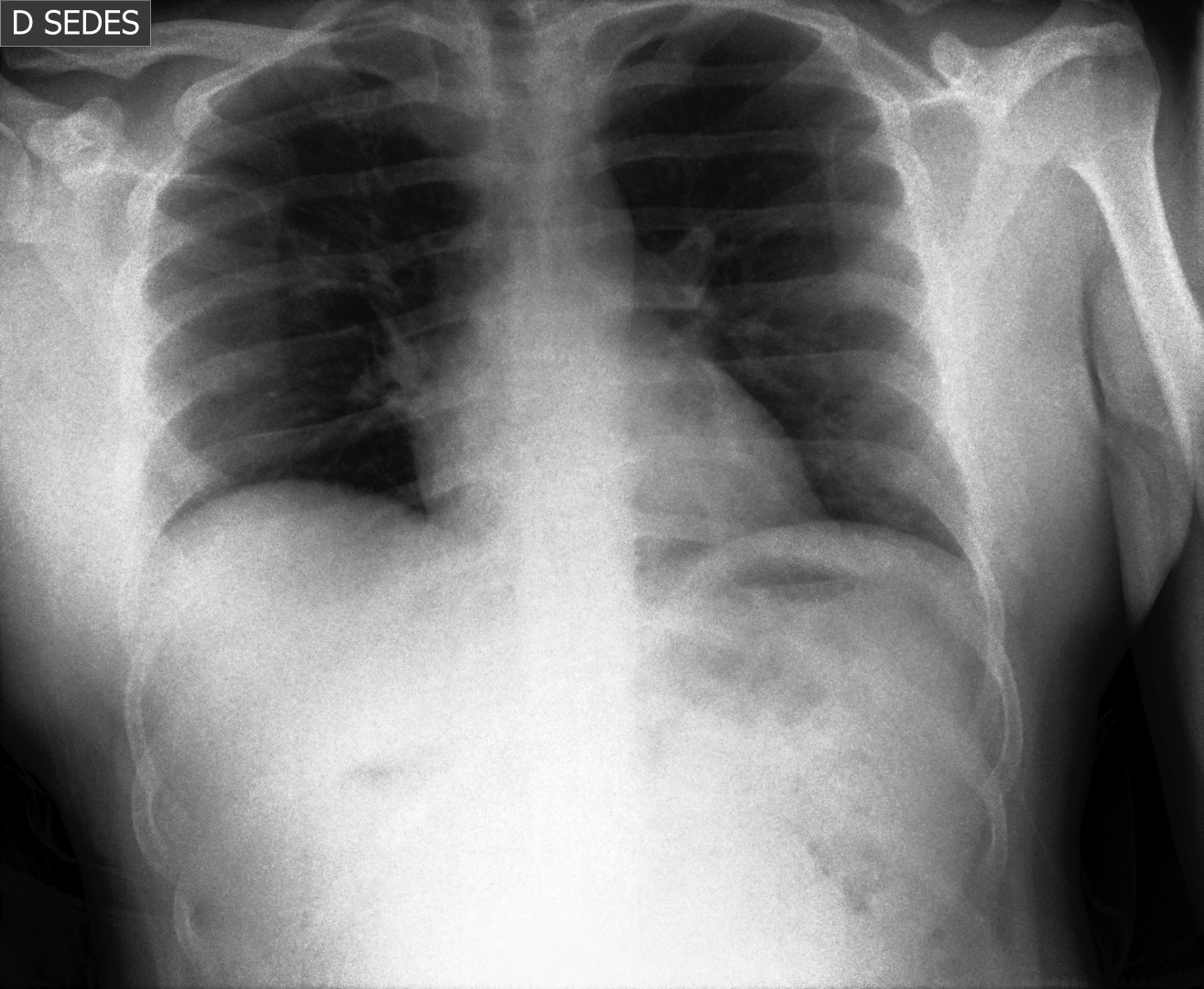}%
    	\label{subfigure:cov1}}
        \hfil
        \subfloat[]{\includegraphics[width=0.2\textwidth]{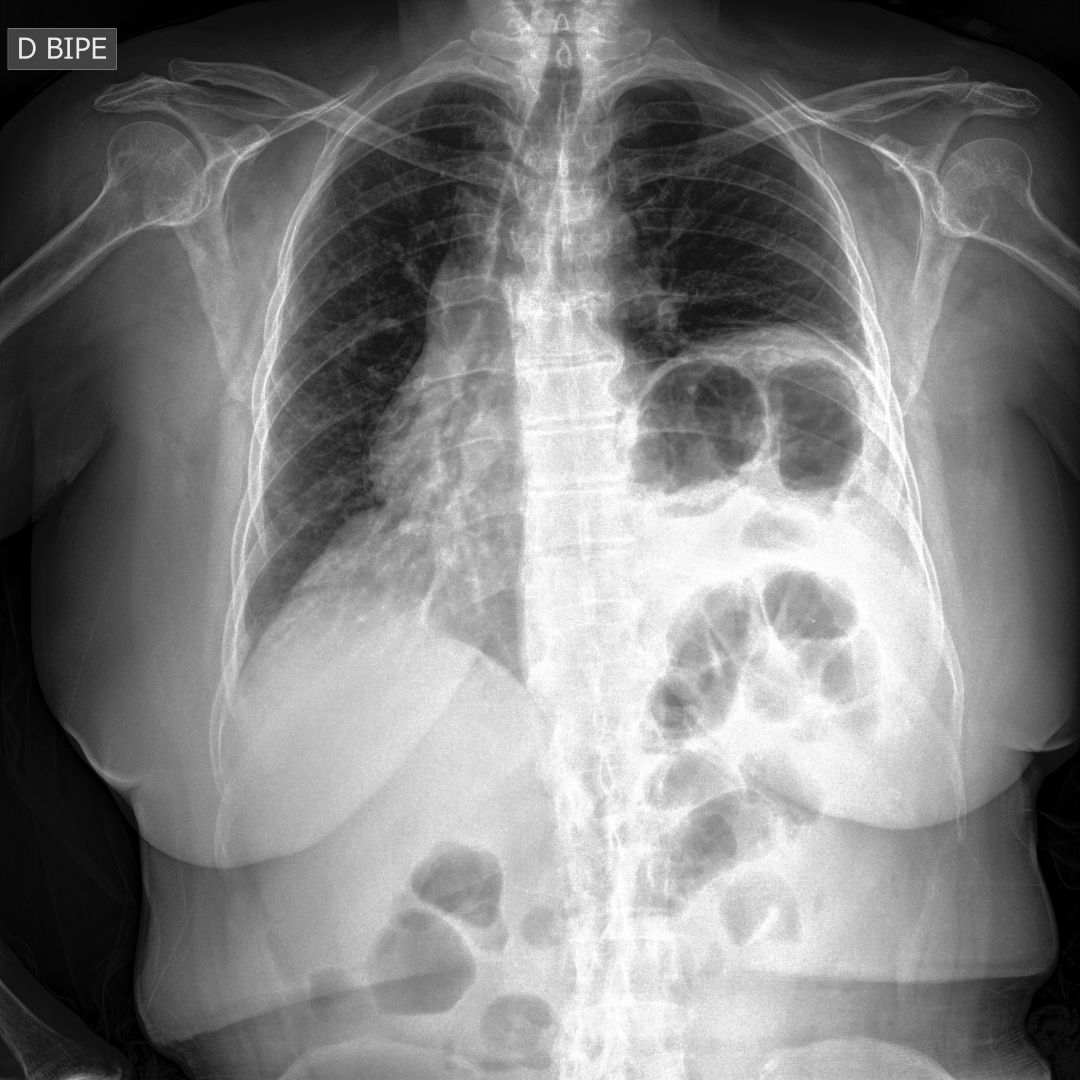}%
    	\label{subfigure:cov2}}
        \hfil
        \subfloat[]{\includegraphics[width=0.2\textwidth]{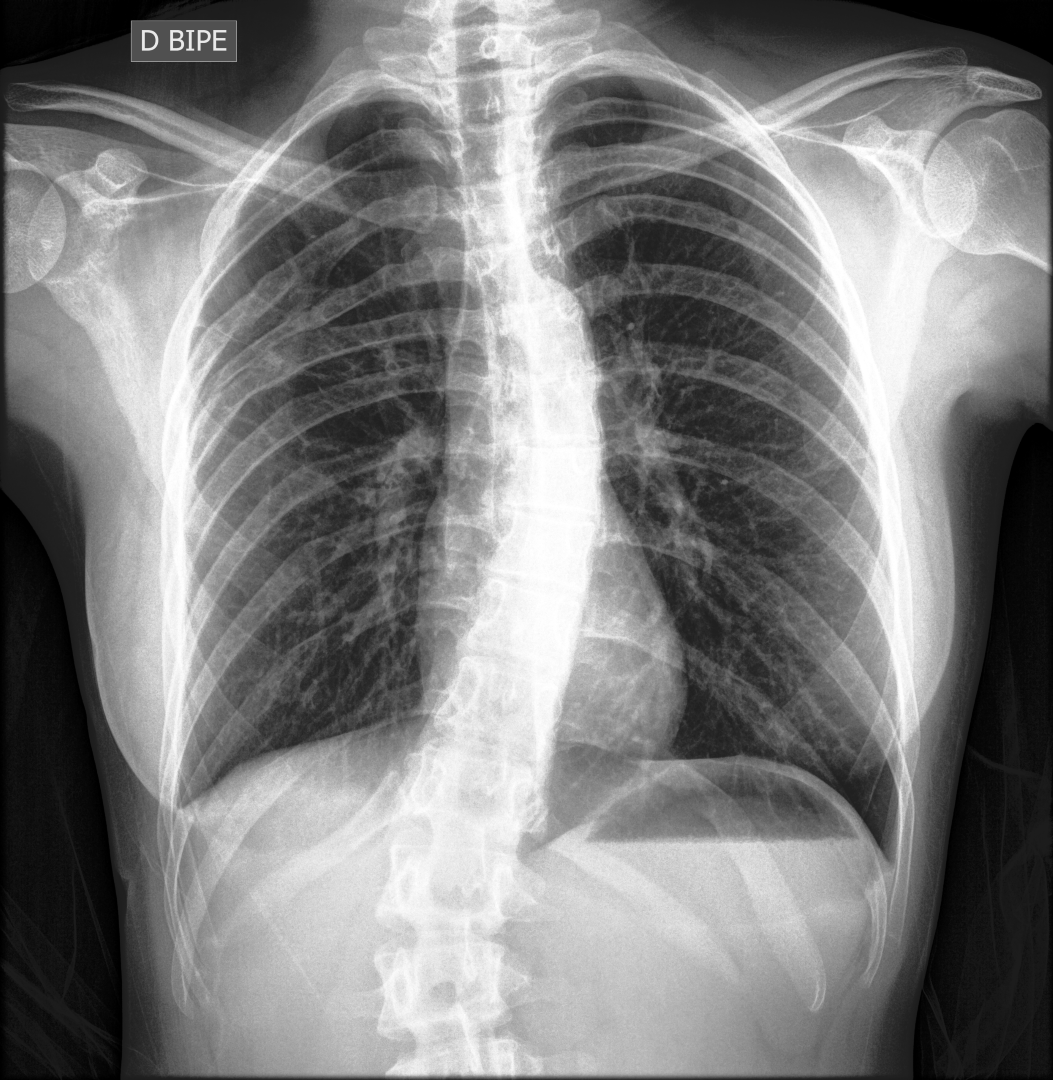}%
    	\label{subfigure:ncov1}}
        \hfil
        \subfloat[]{\includegraphics[width=0.2\textwidth]{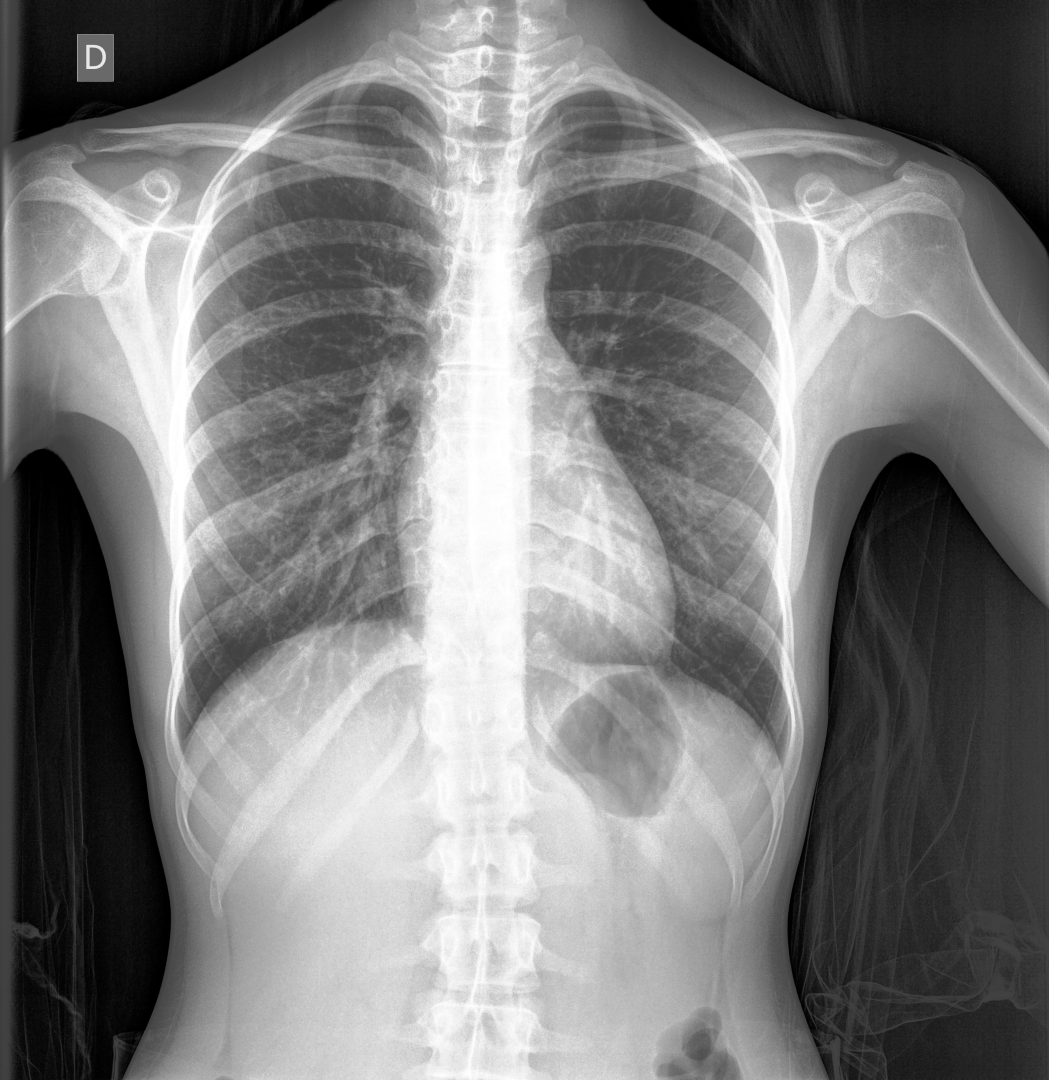}%
    	\label{subfigure:ncov2}}
    	\\
 	\caption{Samples from HUSE dataset: chest x-ray image from different patients with pulmonary involvement (\ref{subfigure:cov1}, \ref{subfigure:cov2}) and without (\ref{subfigure:ncov1}, \ref{subfigure:ncov2}).}
	\label{figure:cmp_v1}
\end{figure}

The image dataset we utilised in this investigation was provided by the University Hospital Son Espases (HUSE) situated in Palma, Spain. In total, 2040 chest x-ray images from patients with and without COVID-19 pneumonia. In Figure \ref{figure:cmp_v1}, we can see samples from this dataset. This dataset and experimentation has been authorised by the Research commission from HUSE (Hospital Universitari Son Espases) (Ref: 3959).  We used as an AI model a ResNet18~\cite{he2016deep}, a well-known DL model for classification of images. { Arias-Duart \emph{et al.}~\cite{arias2022focus} proposed an objective benchmark for \textit{post-hoc} XAI method and identified GradCAM~\cite{selvaraju2017grad} as ``consistently reliable'' in contrast with others largely used XAI techniques. We used this method based on these authors results. } The performance of the trained model can be seen in Table \ref{tab:ch5_train_res}. We consider that the training details are outsided of the scope of this article, centrered on the measurement of trust. Nonetheless, and to allow for a better reproductility we upload the weight of the model in a public repository\footnote{\url{https://github.com/explainingAI/xai_trust}}.


\begin{table}[hbtp]
    \centering
    \begin{tabular}{lc}
        \toprule
        TP              & 46    \\
        FP              & 127   \\
        FN              & 11    \\
        TN              & 490   \\
        Precision       & 0.266 \\
        Recall          & 0.807 \\
        F1-Score        & 0.4   \\ 
        Accuracy        & 0.8 \\
        \bottomrule
    \end{tabular}
    \caption{Metrics obtained with the AI model used in our experimentation.}\label{tab:ch5_train_res}
\end{table}

\begin{figure}[h!]
	\centering
    	\subfloat[Saliency map]{\includegraphics[width=0.19\textwidth]{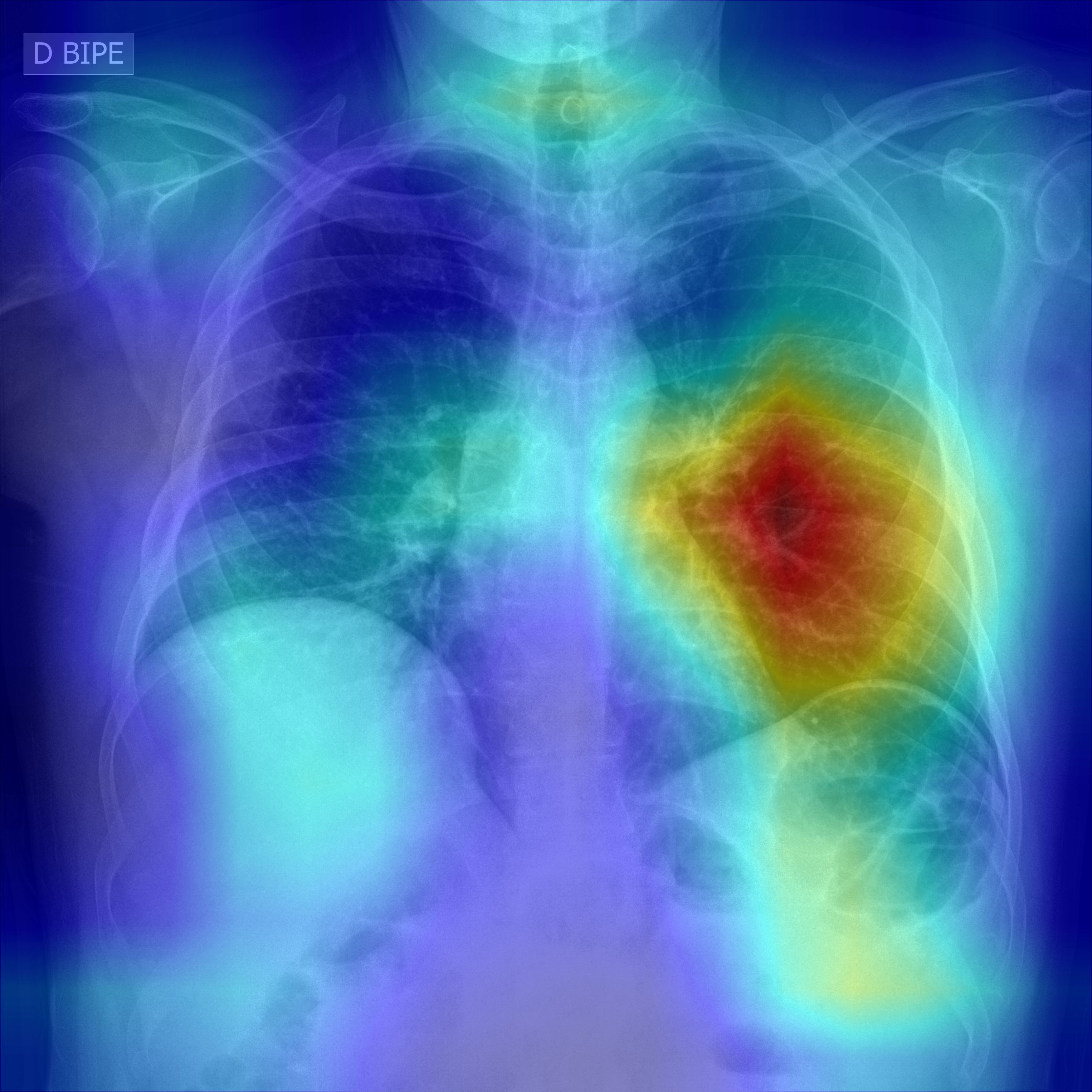}%
    	\label{subfigure:si_si}}
        \hfill
        \subfloat[Threshold of $0.25$]{\includegraphics[width=0.19\textwidth]{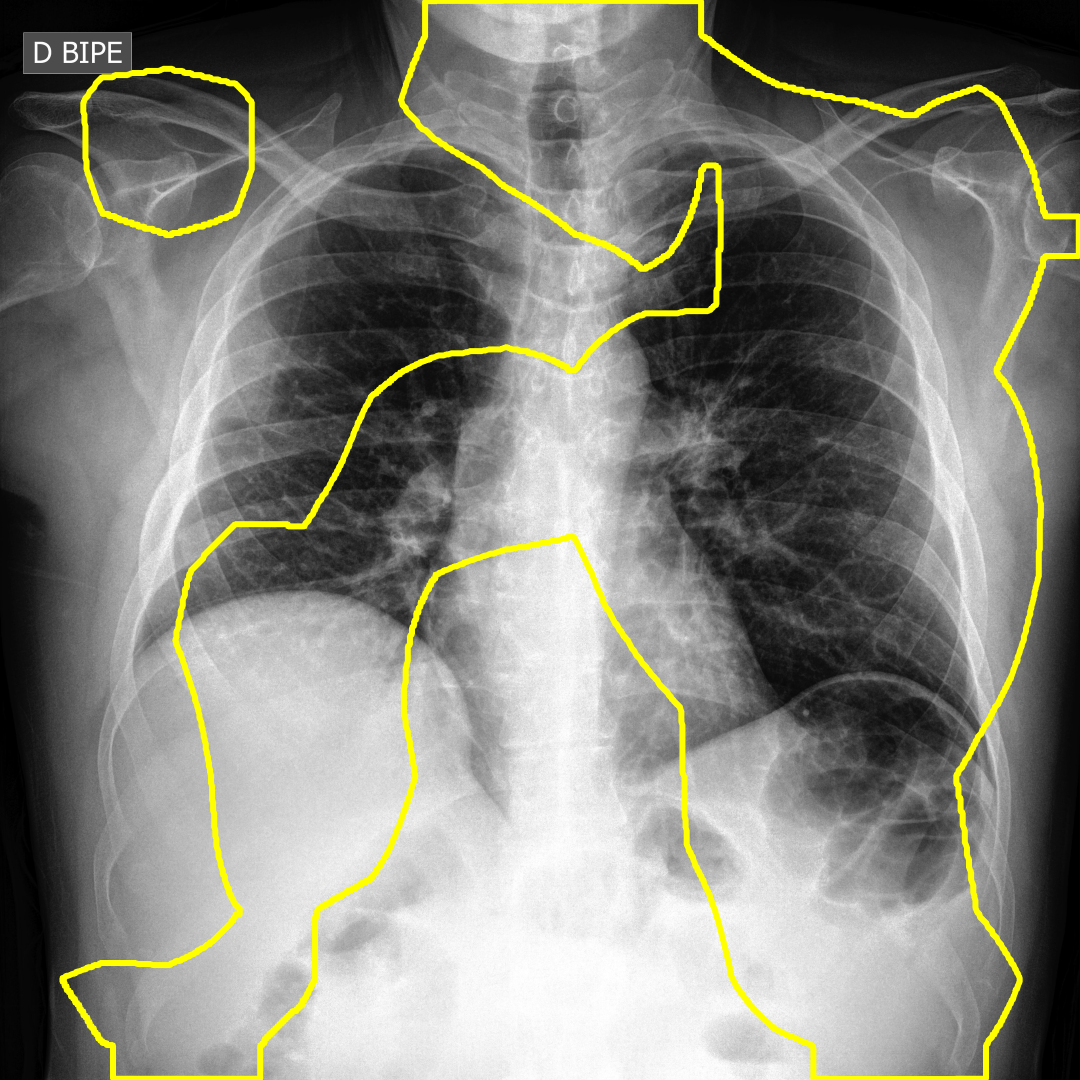}%
    	\label{subfigure:thr_25}}
        \hfill
        \subfloat[Threshold of $0.5$]{\includegraphics[width=0.19\textwidth]{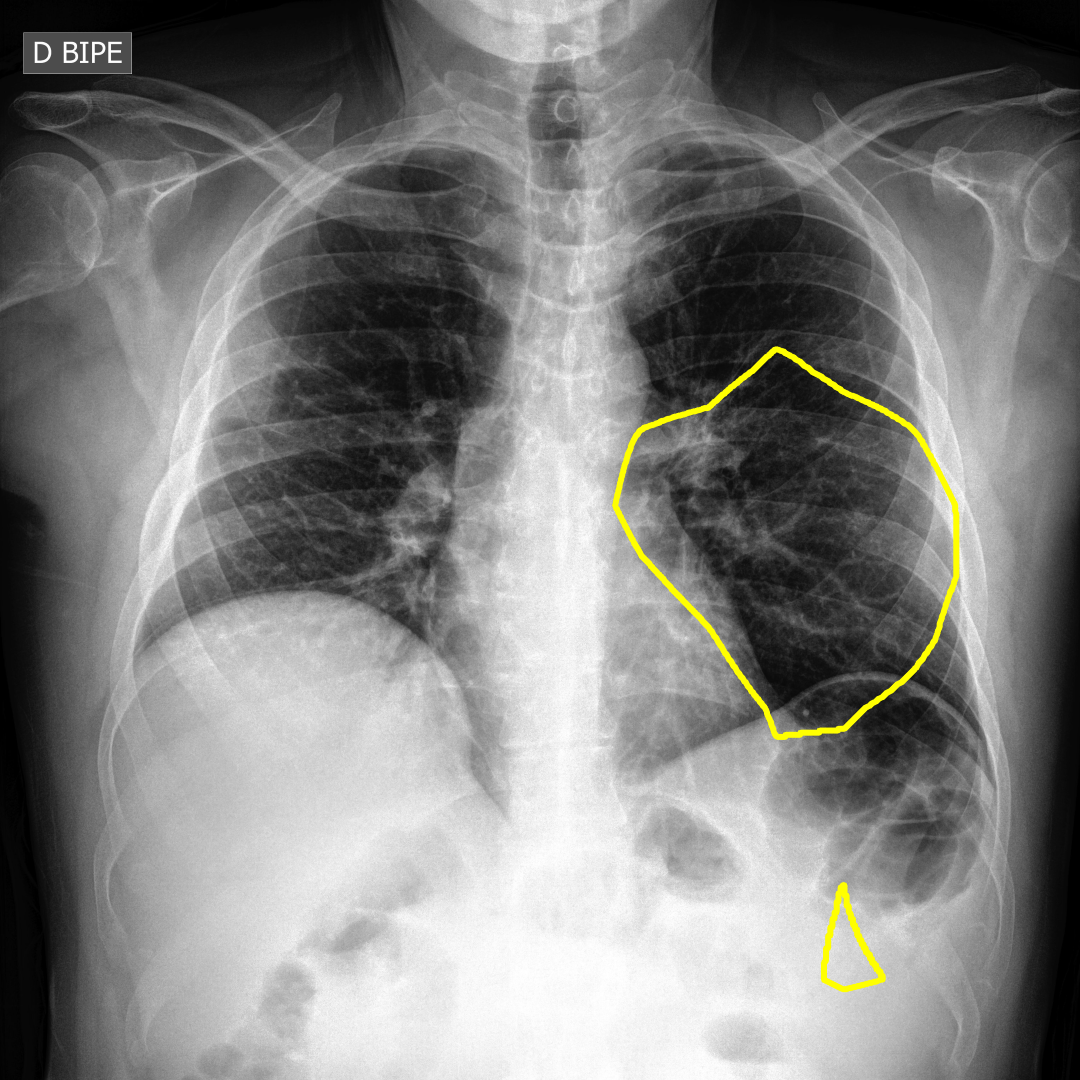}%
    	\label{subfigure:thr_50}}
        \hfill
        \subfloat[Threshold of $0.75$]{\includegraphics[width=0.19\textwidth]{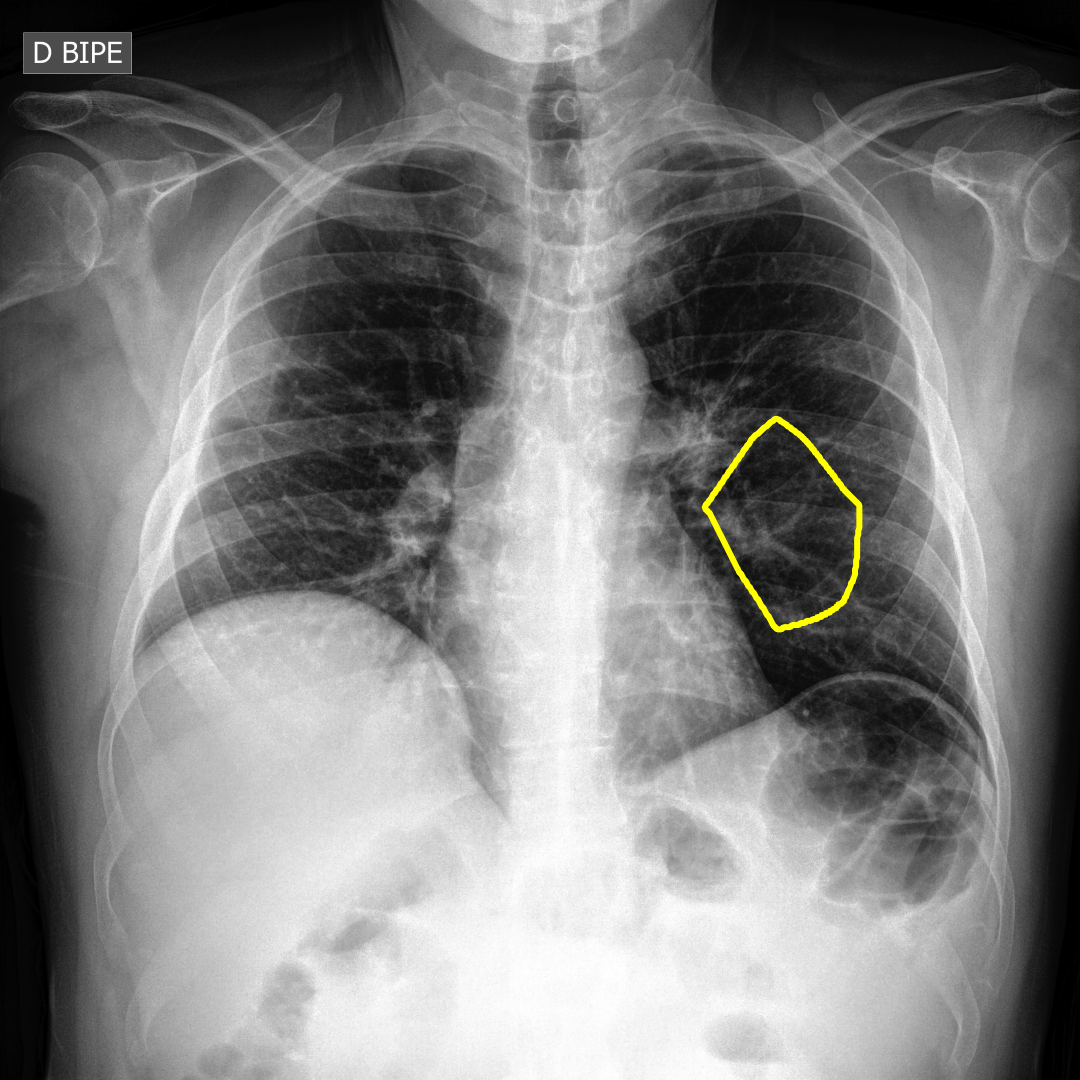}%
    	\label{subfigure:thr_75}}
        \hfill
        \subfloat[Threshold of $0.9$]{\includegraphics[width=0.19\textwidth]{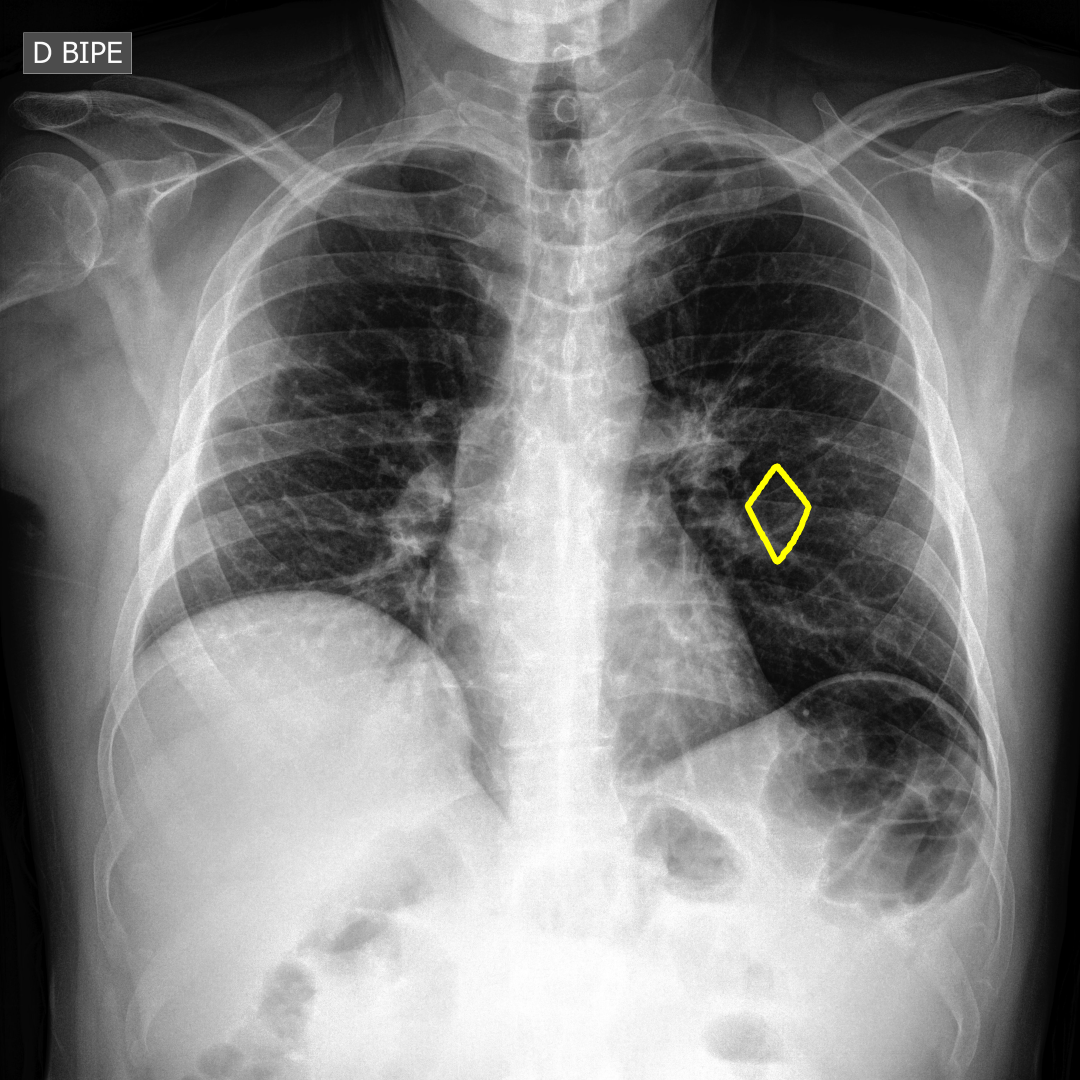}%
    	\label{subfigure:thr_90}}
     \caption{A saliency maps (a) and its simplified views (b,c,d,e), showing all pixels with an importance at least of the threshold indicated below each image.}
	\label{fig:sal_maps}
\end{figure}

 The resulting explanation from GradCAM~\cite{selvaraju2017grad} is a salency map. We consider that, while saliency maps are bastly used, this visualisation can be hard to interpret for a non AI expert user. To simplify them, we highlight the most important parts of the image, we did that using a set of four different thresholds ($0.9$, $0.75$, $0.50$, $0.25$), depicting only the pixels with an importance higher than the threshold. An example of the resulting visualisation can be seen on Figure \ref{fig:sal_maps}. 

\begin{figure}
	\centering        
    	\subfloat[Original Graphical User Interface (GUI) designed.]{\includegraphics[width=0.35\textwidth]{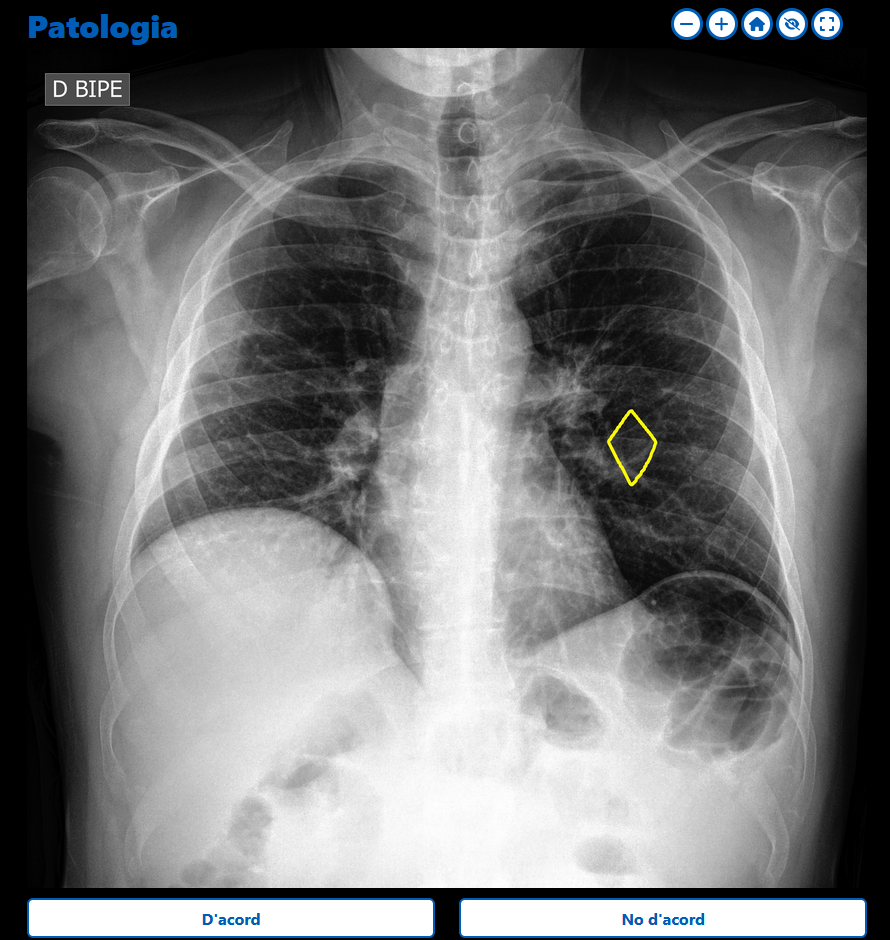}%
    	\label{subfigure:cat_gui}}
        \hspace{5mm}
        \subfloat[Scheme in english of the GUI.]{\includegraphics[width=0.35\textwidth]{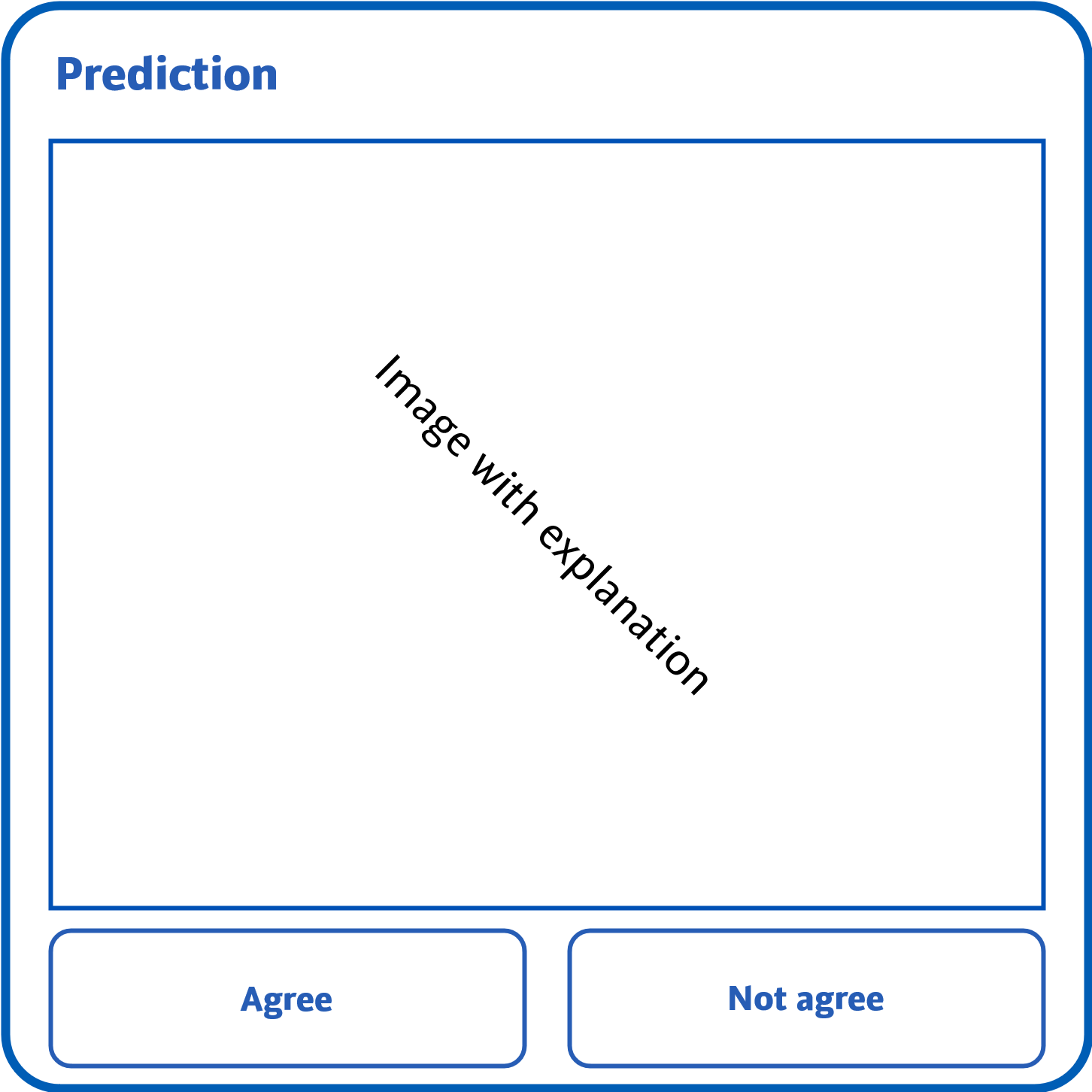}%
    	\label{subfigure:en_gui}}
        \hfill
\caption{Graphical User Interface (GUI) utilised to recollect the user trust. The GUI, originally designed in the doctors language (Catalan) depicts the prediction of the AI model combined with the explanation, and ask the user whether they agree or not to the results. In this case, we only highlight the pixels with an importance higher than $0.9$. In sub-figure \ref{subfigure:en_gui} we can see a schematic version of the GUI in English.}\label{fig:gui}
\end{figure}

Finally, we engaged two highly experienced radiologists as users to measure their trust with the system. The primary data source was collected via the usage of an interactive Graphical User Interface (GUI). We show to the user an interface containing the information regarding the predictions, the explanation and the x-ray image. The proposed GUI can be seen in Figure \ref{fig:gui}. The whole design of this GUI was fine-tuned taking into consideration the radiologist criterion. The two users examined a total of 120 chest x-ray images. 40 of these 120 were presented to both radiologist, while the remainder were only reviewed by one of them. { The user was asked wether agreed with the combination of explanation and prediction or not.}


\begin{table}[!htb] 
    \centering
    \begin{tabular}{lcccc}
        \toprule
        Metric                              & \shortstack{Usr 1 \\ (All imgs)}   & \shortstack{Usr 2 \\ (All imgs)} & \shortstack{Usr 1 \\ (Shared imgs)}   & \shortstack{Usr 2 \\ (Shared imgs)}   \\ \midrule
        TT                                  & 7         & 1     & 4         &   0               \\
        UF                                  & 2         & 1     & 2         &   0    \\
        UT                                  & 57        & 63    & 28        &   32     \\
        TF                                  & 14        & 13    & 6         &   8   \\
        Precision                           & 0.109    & 0.016  & 0.125     &   0   \\
        Recall                              & 0.333    & 0.071  & 0.400     &   0   \\
        F1-Score                            & 0.165    & 0.026  & 0.190     &   0   \\ \midrule
        Lai \& Tan~\cite{lai_human_2019}    & 0.263    & 0.175  & 0.150     &   0   \\
        \bottomrule
    \end{tabular}
    \caption{Results obtained from Users 1 and 2 with all images and only with the images that both user have measure.}\label{tab:res_usr}
\end{table}


We show the results separated depending on the user. The results can be seen in Table \ref{tab:res_usr}, these results shows that exist a significant difference between the results of both users, with a F1-Score values of $0.165$ and $0.028$ respectively. However, both of these values were low enough to consider that the user did not trust the explanations. We considered that this small trust values were a result of the trust trajectory, the fact that the system outputs incorrect prediction affects the following trust measure, even for correct predicted ones. Comparing this results with the ones obtained in the rest of case of studies we can see the similary between them and the \textit{suspicious user} defined in the previous sections. Similarly, with the subset of shared images the trust measures showed lack of consensus between both users, even with the same images. This different results, from users with similar background, indicated the user-dependent nature of any Trust measurement.

{ By analysing the results obtained by Lai \& Tan~\cite{lai_human_2019} and comparing them with our proposal, we observe that both approaches yield relatively low levels of trust. However, their method reports slightly higher trust levels than ours. This difference arises primarily because their approach penalises instances where the user did not trust incorrect predictions.}



In Figure \ref{fig:separated_results} we can see the results obtained for all three metrics with different threshold values. From these three plots, we can see the difference between both users, with an overall higher trust of the first user than the second. We can also see that there were a decrease in the trust when more pixels, including less important ones, were shown to the users (see Figure \ref{fig:sal_maps} for examples of different visualisations). 

\begin{figure}[!htb]
\begin{tikzpicture}
\begin{axis}[
    xlabel={Threshold},
    ylabel={F1 Score},
    symbolic x coords={$>0.25$, $>0.50$, $>0.75$, $>0.90$},
    xtick=data,
    ymin=0, ymax=1,
    legend style={at={(0.95,0.95)},anchor=north east},
    width=7cm, height=5cm,
    grid=both,
]
\addplot[color=blue, mark=*] coordinates {
    ($>0.25$, 0.1)
    ($>0.50$, 0.181818)
    ($>0.75$, 0.190476)
    ($>0.90$, 0.181818)
};
\addlegendentry{User 1}

\addplot[color=red, mark=square*] coordinates {
    ($>0.25$, 0.0)
    ($>0.50$, 0.0)
    ($>0.75$, 0.0)
    ($>0.90$, 0.027020)
};
\addlegendentry{User 2}

\end{axis}
\begin{axis}[
    name=precisionplot,
    xlabel={Threshold},
    ylabel={Precision},
    symbolic x coords={$0.25$, $0.50$, $0.75$, $0.90$},
    xtick=data,
    ymin=0, ymax=1,
    legend style={at={(0.95,0.95)},anchor=north east},
    width=7cm, height=5cm,
    grid=both,
    at={(7.5cm,0cm)},
]

\addplot[color=blue, mark=*] coordinates {
    ($0.25$, 0.0625)
    ($0.50$, 0.125)
    ($0.75$, 0.125)
    ($0.90$, 0.125)
};
\addlegendentry{User 1}

\addplot[color=red, mark=square*] coordinates {
    ($0.25$, 0.0)
    ($0.50$, 0.0)
    ($0.75$, 0.0)
    ($0.90$, 0.0166)
};
\addlegendentry{User 2}
\end{axis}

\begin{axis}[
    name=recallplot,
    xlabel={Threshold},
    ylabel={Recall},
    symbolic x coords={$0.25$, $0.50$, $0.75$, $0.90$},
    xtick=data,
    ymin=0, ymax=1,
    legend style={at={(0.95,0.95)},anchor=north east},
    width=7cm, height=5cm,
    grid=both,
    at={(3.5cm,-6cm)}, 
]

\addplot[color=blue, mark=*] coordinates {
    ($0.25$, 0.25)
    ($0.50$, 0.33)
    ($0.75$, 0.40)
    ($0.90$, 0.33)
};
\addlegendentry{User 1}
\addplot[color=red, mark=square*] coordinates {
    ($0.25$, 0.0)
    ($0.50$, 0.0)
    ($0.75$, 0.0)
    ($0.90$, 0.10)
};
\addlegendentry{User 2}
\end{axis}

\end{tikzpicture}
\caption{Plots showing trust metrics with different threshold values.}\label{fig:separated_results}
\end{figure}



{ To understand the reason behind this lack of trust and, therefore, verifying our proposal, we created a questionnaire to be answered by each user. We asked three distinct questions, each one with different goals, in order to determine the source of the lack of trust in the system. Each question had an attached image to be analysed by the users. Both the questions and the answers can be seen on the \ref{app:question_res}.}

The findings of the questionnaire indicate that the radiologists did not trust the system, therefore, showing the sensible nature of trust to the AI system. The questionnaire responses and conclusions were congruent with the metric results. Both user showed a lack of trust in either the prediction or the explanation, demonstrating the capacity of the proposed measures to assess user trust in a XAI system.

This case study depict both benefits and perks of our proposal. On one hand, identified objectively a lack of trust of both users in the system, even more, quantified this amount. All of that with a behavioural approach to trust, as discussed by Scharowski~\cite{scharowski_trust_2022}, allowing an objective measurement and avoiding the usage of attitudinal (subjective) measures of trust as questionnaires. Our proposal surpassed the previous proposals, mainly the work from Lai \& Tan~\cite{lai_human_2019}, maintaining their simplicity to analyse the results but at the same time with an improved capacity to identify possible problems and allowing a more granular analysis of the results, thanks to the addition of performance information. { We verified this increased granularity with a } further analysis of data recollection in a form of a posterior questionnaire.

\section{Conclusion}
\label{sec:conclusion}

In this study, we proposed a novel trust measure from a behavioural point of view the trust of a user within a XAI system. Our proposal aimed to combine both information about the performance of the predictive model (an objective feature) and the user confidence on the explanation (a subjective feature), allowing a straight forward interpretation of the results, aiming to quantify the system performance effect over the trust in it. In particular, we proposed, to combine classification metrics (True Positives, False Positives, False Negatives, and True Negatives), obtained from the objective comparison between the ground truth and the prediction, and the choice of a user between to trust or not an explanation. The results of this combination were the following measures: Trust True (TT), Untrust True (UT), Trust False (TF), Untrust False (UF). 

These four metrics were based on well-known concepts from social science, such as trust trajectory and the algorithmic aversion. Additionally, from these four metrics, we can use any of the existing high level measure, to get an easy-to-understand result. The key advantage of our solution is the ease in which the data may be analysed. In contrast to other existing trust measures, which are based on questionnaires and have multiscale values, we used non-subjective results, allowing for a simpler and straight forward interpretation. This objective evaluation is a large improvement on the existing literature, as discussed  by Miller \cite{miller2019explanation}, until now most of the evaluation on a XAI context is depending "on the researcher intuitions of what constitutes a 'good' explanation". Furthermore, the increase in granularity, in comparison to other objective proposals to measure trust (Lai \& Tan \cite{lai_human_2019}), allowed us to identify the reason behind the lack of trust.

We defined 3 different case of study to test our proposed measures. The first one is defined by both hypothetical performance and hypothetical trust results. This first case of study allowed us to check the sensitivity of our proposal in three extreme cases: \textit{perfect user}, \textit{entrusted user}, and \textit{suspicious user}. Secondly we test this same hypothetical trust values with a real machine learning model proposed by Petrović \emph{et al.}~\cite{PETROVIC2020104027}. Similarly to the first one, this case of study showed the limitation of the previous work (Lai \& Tan \cite{lai_human_2019}) and the increased ability of our proposal to detect different trust behaviours. { Additionally, this allowed us to verify that our measures are sufficiently sensitive to compare the performance of different models under the same trust levels and to identify distinct behavioural patterns. This suggests a potential future application of our measure: comparing different machine learning models with similar users.} Finally, we defined a third case study, with real performance and trust values. This case of study is based on a XAI pipeline to detect COVID-19 in chest x-ray images. We tested the trust, via a Graphical User Interface, of two radiologists in this pipeline. The results of the metrics indicated an overall lack of trust in the system. 

As future work, we propose defining additional case studies to further explore the relationship between trust and algorithmic aversion. Specifically, the use of a placebo could help assess whether a user's lack of trust persists when they believe that a human-generated outcome is produced by an AI model. This approach would provide deeper insights into the psychological factors influencing trust in AI-driven decisions.

\section*{Acknowledgements}

This study is part of the Project PID2023-149079OBI00 funded by MICIU/AEI/10.13039/501100011033 and by ERDF/EU. The work of Maria Gemma Sempere and Manuel González-Hidalgo was partially supported by the R+D+i Project PID2020-113870GB-I00-``Desarrollo de herramientas de Soft Computing para la Ayuda al Diagnóstico Clínico y a la Gestión de Emergencias (HESOCODICE)'' funded by MICIU/AEI/10.13039/501100011033/.

\bibliographystyle{abbrvnat} 
\bibliography{bibliography}

\appendix
\section{Questionnaire \& results} \label{app:question_res}
\FloatBarrier
\begin{table}
    \centering
        \begin{tabular}{cl}
        \toprule
        Question ID     & Question                                                                                  \\ \midrule
        Q1              & \makecell[l]{This image has been analysed by a radiologist and diagnosed \\ as pathology, do you agree?}   \\
        Q2              & This image contains any pathology?                                                        \\
        Q3              & \makecell[l]{This image has been analysed by a radiologist and diagnosed \\ as healthy, do you agree?}     \\
        \bottomrule
        \end{tabular}
    \caption{Questions used in the questionnaire.}\label{tab:questions}    
\end{table}

\begin{table}
    \centering
        \begin{tabular}{ccllcccc}
        \toprule
        ID  &   Q. ID           & User 1    & User 2    & AI Pred.  & GT    & \makecell{Usr 1 \\ trusted the \\ image?} & \makecell{Usr 2 \\ trusted the \\ image?}   \\ \midrule
        A1  &   Q1              & Yes       & -         & C         & C     & -                          & Yes                   \\
        A2  &   Q2              & Yes       & Yes       & C         & C     & No                         & -                     \\
        A3  &   Q3              & Yes       & Yes       & NC        & NC    & No                         & -                     \\
        A4  &   Q3              & Yes       & Yes       & NC        & NC    & Yes                        & No                    \\
        A5  &   Q3              & No        & -         & C         & C     & No                         & -                     \\
        A6  &   Q1              & No        & -         & C         & NC    & -                          & No                    \\
        A7  &   Q3              & -         & Yes       & C         & NC    & -                          & No                    \\
        A8  &   Q1              & -         & Yes       & C         & C     & Yes                        & -                     \\
        A9  &   Q1              & -         & No        & C         & C     & Yes                        & No                    \\ \bottomrule
        \end{tabular}
    \caption{Answers of the questionnaire from both users. ID identifies the answer for further analysis. Q. ID indicated the question, following the questions defined in Table \ref{tab:questions}. Columns User 1 and User 2, indicate whether the user trusted the explanation originally. AI Pred. showed the prediction of the AI and GT the ground truth. The results are indicated with a C and NC: C is a COVID-19 prediction and NC a healthy prediction. The dashes indicated that the user did not answer that question.}\label{tab:questionnaire}    
\end{table}

Answer A2 allowed us to see that there is a consensus from both users and the GT with the diagnosis of pathological patient from the image. However, User 1 did not agree with the results and explanation. This fact revealed that a portion of the mistrust was generated {either by the trust trajectory or } by the explanations rather than AI performance, in consequence the user could not trust a right classification. Similarly, we wanted to determine why an inaccurate classification was not trusted as the A6 solution. We can see that the lack of trust was strongly associated to the miss-classification, indicating that the user will not accept an incorrect AI prediction (specified by User 1, User 2, and the Ground truth), showing the relation between trust and the classification performance, one of the basic assumptions from our proposal.

\end{document}